%% file: main.tex
\documentclass[10pt,twocolumn,letterpaper]{article}

\usepackage{amssymb}
\usepackage[pagenumbers]{cvpr} 
\usepackage{tabularx} 
\usepackage{float}

\input{preamble}

\definecolor{cvprblue}{rgb}{0.21,0.49,0.74}
\usepackage[pagebackref,breaklinks,colorlinks,citecolor=cvprblue]{hyperref}
\usepackage{multirow}
\usepackage{graphicx}

\def\paperID{6655} 
\def\confName{CVPR}
\def\confYear{2024}

\DeclareMathAlphabet\mathbfcal{OMS}{cmsy}{b}{n}

\title{FlashVideo: A Framework for Swift Inference in Text-to-Video Generation}

\author{Bin Lei\\
University of Connecticut\\
USA, CT, Storr\\
{\tt\small bin.lei@uconn.edu}
\and
Le Chen\\
Iowa State University\\
USA, IA, Ames\\
{\tt\small lechen@iastate.edu}
\and
Caiwen Ding\\
University of Connecticut\\
USA, CT, Storr\\
{\tt\small caiwen.ding@uconn.edu }
}

\begin{document}
\twocolumn[{%
\renewcommand\twocolumn[1][]{#1}%
\maketitle
\begin{center}
    \centering
    \captionsetup{type=figure}

    \includegraphics[width=\textwidth]{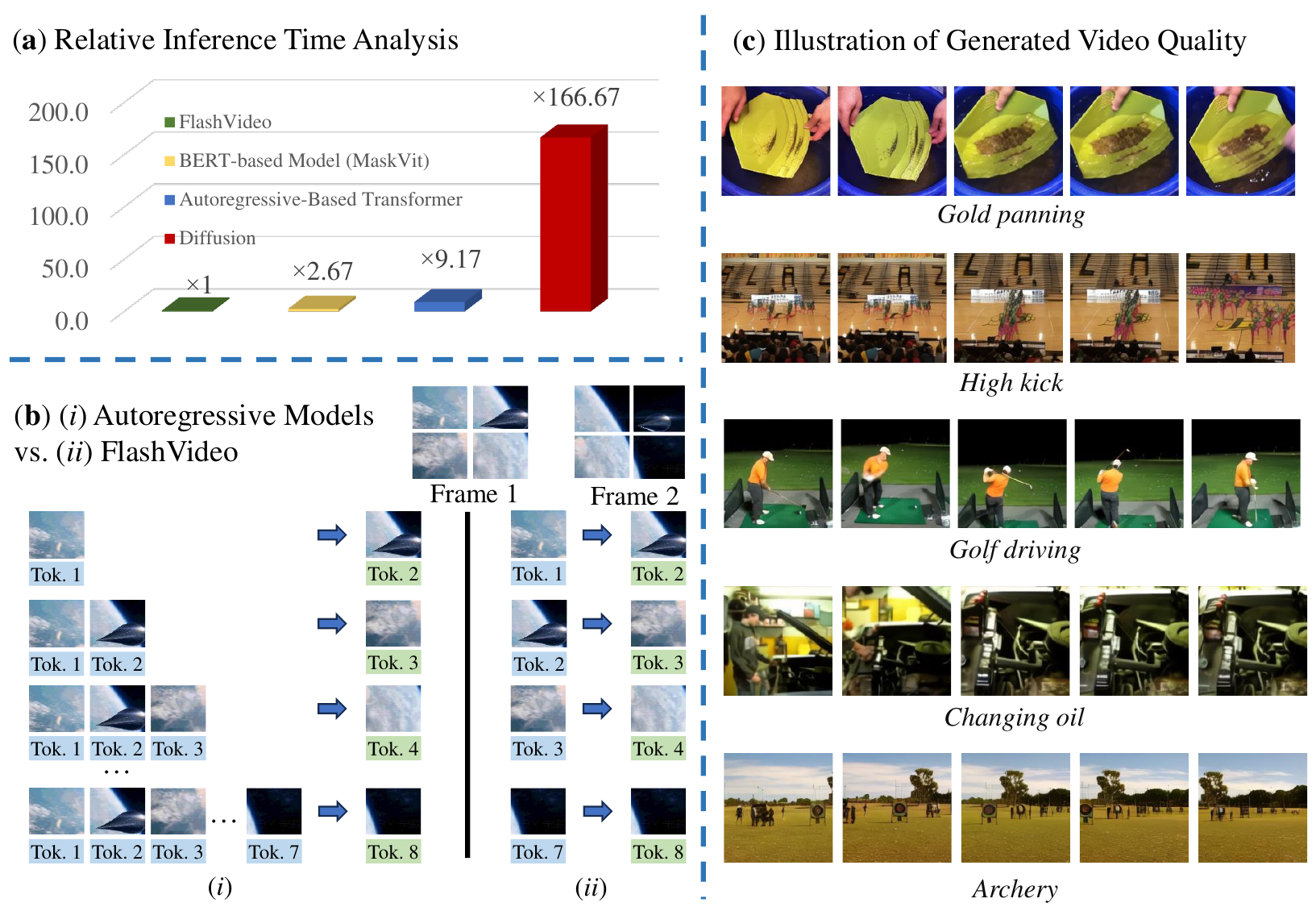}
    
    \captionof{figure}{
Overview of FlashVideo's Video Generation (a) Efficiency, (b) Comparison of the vision token generation methods between Autoregressive Models and FlashVideo, and (c) Quality. 
 (a) compares the relative time taken to generate a single frame by various methods. In (b), we illustrate the reasons behind the increased efficiency of our method compared to the painful slowness of autoregressive-based transformers. (c) displays some of the frames generated by our model, showcasing the quality of the video output. 
}    \label{fig:overview_big}
\end{center}%
}]

\input{sec/0_abstract}


\input{sec/1_introduction}

\input{sec/2_Related_Work}

\input{sec/3_Methology}

\input{sec/4_Experiment}
\input{sec/5_Conclusion}
{
    \small
    \bibliographystyle{ieeenat_fullname}
    \bibliography{main}
}


\end{document}

%% file: preamble.tex
\usepackage[dvipsnames]{xcolor}


%% file: sec/0_abstract.tex
\begin{abstract}
In the evolving field of machine learning, video generation has witnessed significant advancements with autoregressive-based transformer models and diffusion models, known for synthesizing dynamic and realistic scenes. However, these models often face challenges with prolonged inference times, even for generating short video clips such as GIFs. This paper introduces FlashVideo, a novel framework tailored for swift Text-to-Video generation. FlashVideo represents the first successful adaptation of the RetNet architecture for video generation, bringing a unique approach to the field. Leveraging the RetNet-based architecture, FlashVideo reduces the time complexity of inference from $\mathcal{O}(L^2)$ to $\mathcal{O}(L)$ for a sequence of length $L$, significantly accelerating inference speed. Additionally, we adopt a redundant-free frame interpolation method, enhancing the efficiency of frame interpolation. Our comprehensive experiments demonstrate that FlashVideo achieves a $\times9.17$ efficiency improvement over a traditional autoregressive-based transformer model, and its inference speed is of the same order of magnitude as that of BERT-based transformer models. 
\end{abstract}

%% file: sec/1_introduction.tex
\section{Introduction}

Despite the availability of mature frameworks for video generation tasks, such as Generative Adversarial Networks (GANs)~\cite{creswell2018generative,clark2019adversarial,vondrick2016generating}, Transformer-based models~\cite{ranftl2021vision,villegas2022phenaki,rakhimov2020latent}, and diffusion models~\cite{Ho2022VideoDM,voleti2022mcvd,gu2022vector}, each presents distinct strengths and limitations.  GANs, a cornerstone in generative modeling particularly for images, face notable challenges in video generation such as maintaining temporal coherence and consistency across frames, and high computational demands for capturing long-term dependencies. Transformer-based models, adept at handling long-range dependencies, mitigate some GAN limitations but encounter increased inference times when processing the complex, multi-frame structure of videos. Diffusion models, a more recent development primarily in image generation, demonstrate prowess in generating high-fidelity outputs but are hampered by slow inference speeds, a significant hurdle when extended to the intricate domain of video generation.


In the ever-evolving landscape of machine learning, particularly within natural language processing (NLP), there has been a push towards more efficient reference architectures. A notable advancement in this direction is the introduction of RetNet by \citet{sun2023retentive}, poised as a potential “successor to the Transformer”. RetNet's innovative architecture, blending parallel computation with recurrent processing, has garnered considerable attention. Its hybrid design facilitates rapid training, akin to traditional Transformers, yet efficiently handles extensive datasets. Importantly, during inference, RetNet adopts a recurrent mode, significantly reducing the sequence length's impact on processing time. This feature is invaluable for tasks with long sequences where computational demands escalate with each added element. In the realm of video generation, our model, FlashVideo, harnesses RetNet's architecture to enhance frame generation efficiency. As illustrated in Figure~\ref{fig:overview_big}~(b), unlike traditional autoregressive transformer-based models that generate the next frame based on a sequence of previous frames, FlashVideo innovatively generates each frame primarily from its immediate predecessor. This approach, depicted in Figure ~\ref{fig:overview_big}~(a), markedly boosts inference efficiency, a crucial advantage in video processing.


Adapting RetNet for video generation tasks presents significant challenges, especially considering its recent introduction and the absence of prior applications in this area. This unexplored territory in video generation with RetNet poses unique technical hurdles.
The biggest challenge lies in the implementation of an effective attention mechanism that operates both across and within video frames. Contrasting with approaches like CogVideo~\cite{hong2022cogvideo}, which effectively segregates temporal and spatial attention, adapting RetNet in a similar manner is complex. The crux of this challenge stems from RetNet's use of relative position encoding. This differs fundamentally from the absolute position encoding utilized in traditional transformers, where positions within frames are explicitly encoded to facilitate intra-frame attention. With RetNet's relative positioning, re-encoding these positions to compute inter-frame attention becomes a non-trivial task, thus complicating the adaptation of RetNet for local frame attention requirements in video generation.

Due to RetNet's use of relative position encoding, we are unable to employ the traditional bifurcated channel technique, akin to the one used in CogVideo, which separates models into temporal and spatial channels to discern which attentions belong within frames and which between them. To address this challenge, we have adopted a strategy that incorporates Serial Number tokens. The crux of this approach lies in leveraging textual information to compensate for the partial positional information that is missing. This enables the model to accurately distinguish between intra-frame attention and inter-frame attention, thereby enhancing its understanding of the temporal and spatial context.



In response to the challenges identified in adapting RetNet for video generation, this paper pioneers three key innovations to navigate and overcome these difficulties. Firstly, we develop tailored training and inference frameworks for the RetNet model, specifically for key stages in video generation: key frame generation and frame interpolation. This approach ensures that RetNet is effectively adapted to the unique demands of video content generation. Secondly, we introduce an advanced sequencing technique, designed to enhance RetNet's capability in understanding and learning inter-frame relationships, a critical aspect of maintaining temporal coherence in videos. Lastly, we propose an innovative Redundant-free Frame Interpolation method to enhance the interpolation process's efficiency. This method strategically interpolates only the essential regions of each frame to the video's continuity, thereby optimizing the computational resources and reducing processing time without compromising the video's quality, as shown in Figure ~\ref{fig:overview_big}~(c).

As depicted in Figure~\ref{fig:overview_big}~{(a)}, FlashVideo effectively leverages the RetNet architecture to achieve swift inference in video generation tasks. When compared to traditional autoregressive-based transformer models, FlashVideo realizes an impressive $\times9.17$ efficiency boost, aligning its inference time with that of BERT-based transformer models. This remarkable enhancement not only demonstrates the practicality and effectiveness of FlashVideo but also underscores the significance of our contributions in this domain. Our contributions in this paper are summarized as follows:

\begin{itemize}
\item \textbf{Pioneering adaptation of RetNet for video generation}: This paper marks the first successful adaptation of RetNet, originally an NLP-focused architecture, to the realm of video generation. We address and overcome the unique challenges posed by RetNet's relative position encoding, setting a precedent in the field.
\item \textbf{Tailored training and inference frameworks for video generation}: We innovatively adapt RetNet for video generation by devising specialized training and inference frameworks. Overcoming the limitations of RetNet's relative position encoding, our frameworks enable the effective use of RetNet in video generation, breaking new ground in the application of relative encoding models in this field.
\item \textbf{Innovative redundant-free frame interpolation method}: We propose an effective Redundant-free Frame Interpolation method that maintains high video quality while optimizing computational resources.
\item \textbf{Empirical validation of FlashVideo's efficiency and quality}: Through comprehensive experiments, we demonstrate the efficiency and quality of FlashVideo. These experiments validate our methods and showcase FlashVideo's enhanced performance in video generation tasks.
\end{itemize}

%% file: sec/2_Related_Work.tex
\section{Related Work}
This section introduces the background and related work of video generation and RetNet, giving an overview of the key developments and methodologies that have shaped the field.
\subsection{Video Generation}

The field of video generation has evolved significantly, advancing from traditional deterministic methods to sophisticated generative models capable of synthesizing dynamic, realistic scenes. Early approaches like CDNA~\cite{finn2016unsupervised} and PredRNN~\cite{wang2017predrnn} employed CNNs or RNNs to predict future frames based on initial inputs. However, these methods struggled with capturing stochastic temporal patterns, a challenge later addressed by the advent of Generative Adversarial Networks (GANs)~\cite{goodfellow2020generative}. GANs revolutionized the field by enabling the generation of videos without reliance on initial frames, facilitating both unconditional and class-conditional video synthesis. Various GAN-based models~\cite{bao2017cvae, han2018gan, wang2020imaginator} have been developed for image and video generation.

More recently, the focus has shifted towards text-to-video (T2V) generation, driven by the development of models like VQVAE~\cite{van2017neural} and autoregressive-based transformers~\cite{vaswani2017attention, brown2020language}. These methods have become the mainstream, as seen in works by ~\citet{ho2020denoising}, who proposed a video diffusion model for text-to-video generation. Yet, these methods often face constraints from being trained on specific datasets like UCF-101, leading to domain-specific limitations and a scarcity of publicly accessible models. Innovative contributions like GODIVA~\cite{wu2021godiva} and N\"{U}WA~\cite{wu2022nuwa} have introduced more advanced techniques, including 2D VQVAE with sparse attention and unified multitask learning representations. Building upon these, models such as CogVideo~\cite{hong2022cogvideo} and Video Diffusion Models (VDM)~\cite{Ho2022VideoDM} have integrated temporal attention mechanisms and space-time factorized U-Nets, trained on extensive, privately collected text-video pairs.

Despite these advancements, both transformer-based and diffusion models face the challenge of slow inference due to the necessity of multiple forward and backward network passes. This issue is particularly pronounced in video generation, where processing multiple frames significantly intensifies computational demands.

\subsection{Retentive Network}
The Retentive Network (RetNet) is introduced as a successor transformer architecture for large language models. It distinguishes itself from traditional transformers by incorporating a retention mechanism, which facilitates explicit decay for positional relationship modeling and enables both parallel and recurrent computational modes. RetNet has shown its advantage specifically in inference efficiency (in terms of memory, speed, and latency), favorable training parallelization, and competitive performance. These attributes render RetNet particularly suitable for large language models and video generation tasks, especially considering its $\mathcal{O}(L)$ inference complexity for a sequence with length $L$. In this section, we briefly introduce the key components of RetNet.

\textbf{Parallel Representation}: In this phase, modifies the original transformer model by incorporating relative positional encoding into the query \( Q \)  and key \( K \) computations, while the value \( V \) computation remains unchanged. The specific calculation process can be expressed as:


\begin{align}
Q = (XW_q) \odot \Theta &,  K = (XW_k) \odot \overline{\Theta}, V = XW_v \nonumber \\
\Theta_n = e^{in\theta} &, D_{nm} = 
\begin{cases} 
\gamma^{n-m}, & \text{for } n \geq m, \\
0, & \text{for } n < m,
\end{cases} \nonumber \\
\text{Retention}(X) &= (QK^T \odot D)V. \label{eq:1}
\end{align}
`

In Equation ~\ref{eq:1}, \( \Theta \) represents a method for encoding relative positions in the complex plane. \( \overline{\Theta} \) is the complex conjugate of \( \Theta \). A lower triangular matrix \( D \) ensures that each sequence position only receives information from preceding positions. In RetNet, the information for a given position is derived exclusively from the information of preceding positions. This design ensures that each position in the sequence is informed only by its antecedent elements, adhering to a strict sequential dependency \( \gamma \) is a constant defined by the authors.

\textbf{Recurrent Representation:} The Recurrent Representation of RetNet follows a specific computational process:
\begin{align}
S_n &= \gamma S_{n-1} + K_n^T V_n, \nonumber \\
\text{Retention}(X_n) &= Q_{n}S_{n}, \quad n = 1, \ldots, |x|.
\label{eq:2}
\end{align}
In Equation \ref{eq:2}, \( S \) denotes the hidden state, and it evolves sequentially with each position in the sequence. The variables \( Q \), \( K \), and \( V \) retain their respective roles as in the Parallel representation. A key insight from the original paper on RetNet is the mathematical equivalence of its parallel and recurrent computational forms. This distinctive feature of RetNet allows for efficient parallel training while enabling fast and effective recurrent inference. When compared to traditional Transformer models, especially those with an extensive parameter count (exceeding 2 billion), RetNet demonstrates superior performance. It not only outperforms various iterations of Transformers in language modeling tasks but also shows marked improvements in memory usage efficiency, throughput, and reduced inference latency. These attributes make RetNet particularly advantageous for large-scale applications, such as video generation, where computational efficiency and speed are of paramount importance.



This feature allows the network to be trained in parallel while performing recurrent inference, significantly speeding up the inference process. Compared to Transformer models, particularly those with over 2 billion parameters, RetNet shows superior performance. It surpasses various Transformer iterations in language modeling tasks and demonstrates improved efficiency in memory usage, throughput, and inference latency.

%% file: sec/3_Methology.tex
\section{FlashVideo}
This section introduces the design of FlashVideo. We begin with an overview of FlashVideo, outlining its core architecture and the novel integration of RetNet within this framework. Subsequently, we delve into the specific design strategies implemented to overcome the limitations of RetNet for application in video generation tasks. Following this, we explore our redundant-free frame interpolation method. 
\subsection{Overview}
\begin{figure*}[htbp]
  \centering
  \includegraphics[width=\textwidth]{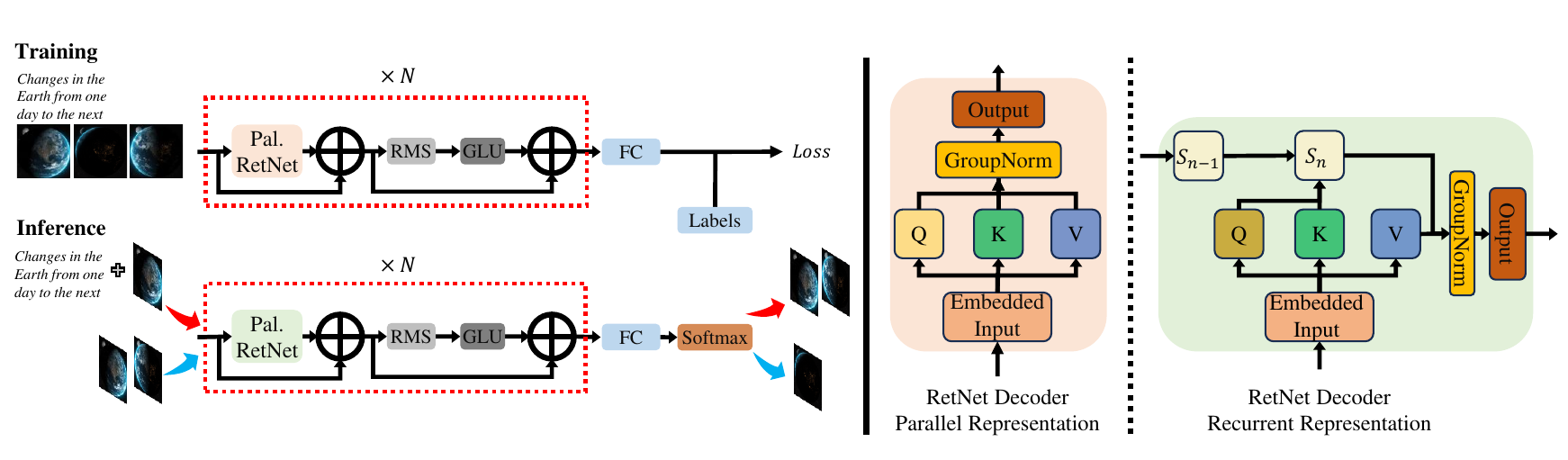}
  \caption{Model Overview. Pal. RetNet: RetNet Decoder Parallel Representation; Rec. RetNet: RetNet Decoder Recurrent Representation; RMS: Root Mean Square Normalization; GLU: Gated Linear Unit activation function; FC: Fully connected layer; $\oplus$: Residual connection; $N$: Number of decoders; \includegraphics[height=1em]{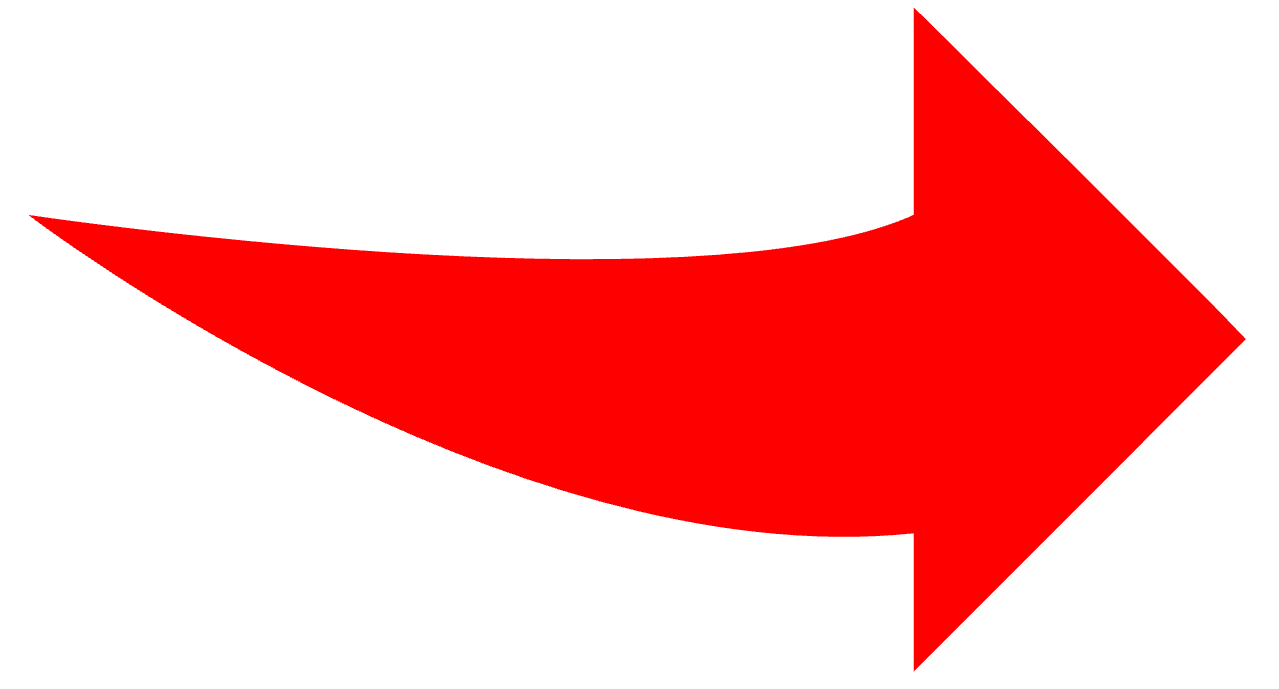}: Input and output for the key frames generation tasks; \includegraphics[height=1em]{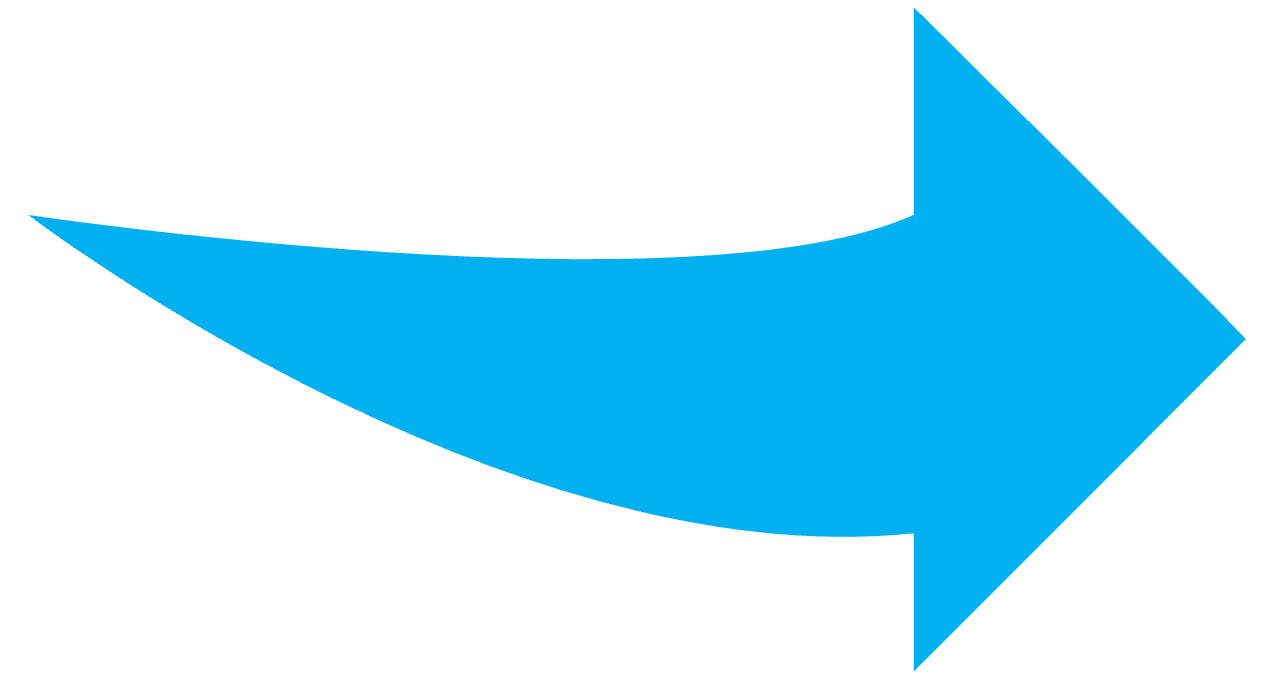}: Input and output for the frames interpolation tasks. The illustration of the RetNet decoder is inspired by their original paper \cite{sun2023retentive}.}
  \label{fig:Model_Overview}
\end{figure*}
Figure~\ref{fig:Model_Overview} presents a comprehensive overview of the FlashVideo model. The figure is divided into two main sections: the left part details the training and inference mechanisms within FlashVideo, while the right side illustrates the specific computational workflows employed in RetNet's parallel and recurrent modes.

During training, FlashVideo ingests textual descriptions alongside multiple frames from the target video. Each frame undergoes segmentation into vision tokens, which are subsequently flattened into a one-dimensional format. Corresponding labels are generated using the teacher forcing~\cite{toomarian1992learning} method. Utilizing RetNet's parallel representation, FlashVideo leverages GPU acceleration in the training phase, ensuring swift training speeds.


In the inference phase, FlashVideo undertakes two primary tasks: generating key frames based on textual input and interpolating frames to construct the video. 
This inference phase process largely mirrors the training structure. The final step utilizes a softmax function to generate a probability distribution over each token, from which selections are made randomly based on these probabilities. In this stage, we employ the recurrent representation method of RetNet, allowing the model to process only the current input token at a time, eliminating the need to handle all previous tokens as is the case with traditional autoregressive transformers. This strategic modification reduces the time complexity of inference from $\mathcal{O}(L^2)$ to $\mathcal{O}(L)$ for a sequence of length $L$, thus significantly boosting inference speed.
 


Furthermore, in FlashVideo, we have incorporated residual connections after both the RetNet blocks and the activation function. These connections, along with the replacement of GroupNorm normalization with RMS normalization and the use of Gated Linear Units (GLU) for non-linear activation, are vital in stabilizing the training process and enhancing overall model performance.




\subsection{Serial Number Token}
\noindent \textbf{Preliminaries.} Addressing the temporal dynamics between frames is a significant challenge in video generation. 
Traditionally, autoregressive-based transformer methods like CogVideo~\cite{hong2022cogvideo} have used separate channels for temporal and spatial attention, handling inter-frame and intra-frame dependencies. 
For example, consider a scenario in a frame comprising 
$n$ vision tokens. In traditional transformer models employing separate time and spatial channels, the position encoding is calculated distinctively for each channel. For the time channel, the position encoding of the first token in the $m$-th frame is computed as 
$m\times n + 1$.  This calculation incorporates both the temporal position of the frame ($m$) and the spatial position of the token within the frame. In contrast, for the spatial channel, the position encoding for the first token remains consistently at 
$1$, irrespective of its temporal placement. This method allows traditional transformers to distinctly encode tokens across time and space, facilitating the model's understanding of both inter-frame (temporal) and intra-frame (spatial) relationships.
Diverging from this, RetNet employs a relative position encoding strategy named Xpos~\cite{sun2022length}. 
This approach makes using separate temporal and spatial channels ineffective for RetNet, as their computational results would be identical, thus limiting traditional dual-channel training methods for learning temporal relationships between frames.

\begin{figure}[htbp]
  \centering
\includegraphics[width=\linewidth]{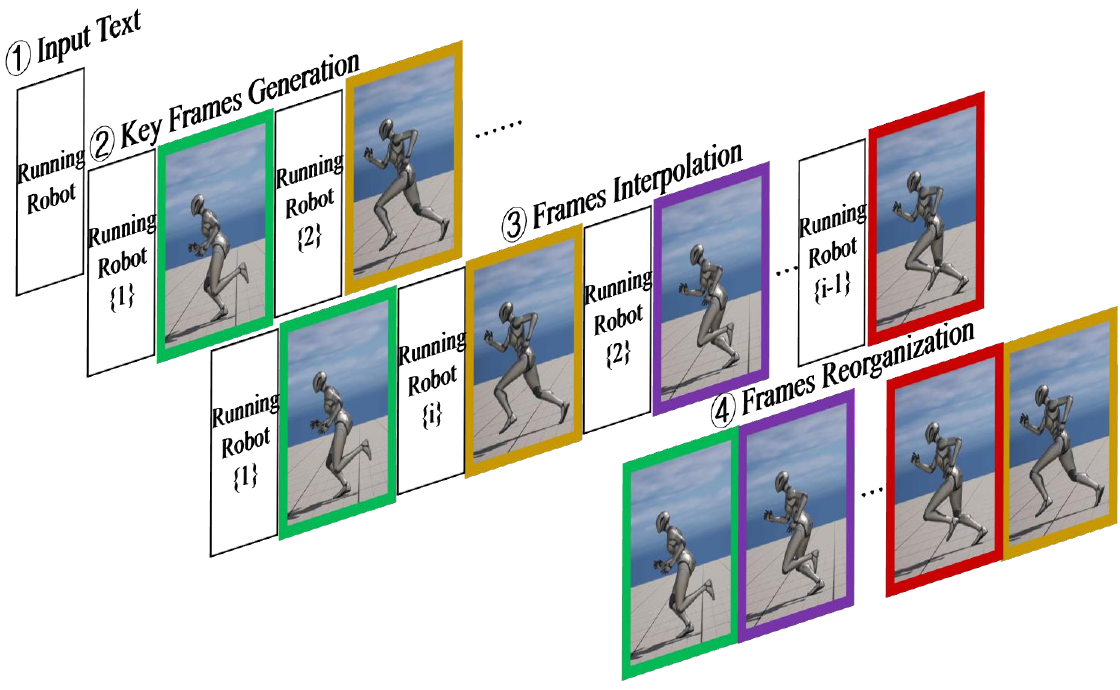}
  \caption{The specific handling of the input text and serial number tokens during key steps in the video generation process. Frames with the same color border represent the same frame.}
\label{fig:Serial_Number}
\end{figure}

\noindent \textbf{Our method.} To enable FlashVideo to learn the relationships between frames, we introduce a novel method by prepending the input text to each image and adding a learnable \texttt{Serial Number} token. This token, combined with the repetitive text input, reinforces positional information with textual context.
This approach is easily applied during the training process, similar to the traditional method of adding a \texttt{Start of Image} special token~\cite{ramesh2021zero} before each frame. In the data preprocessing stage, a class label's text input and its corresponding \texttt{Serial Number} are added before each frame. For the inference stage, the specific processing approach is illustrated in Figure~\ref{fig:Serial_Number}.


The inference process is composed of two main steps: key frame generation and frame interpolation. In key frame generation, the content of the input text is systematically reiterated prior to the construction of each frame, succeeded by adding the \texttt{Serial Number}. This recurrent emphasis on the input text prior to each new frame generation serves to bolster the model's adherence to the textual context, while the \texttt{Serial Number} is instrumental in imparting the model with an awareness of the sequential chronology among frames. The allocation of the \texttt{Serial Number} is intrinsically linked to the index of the imminent frame; for instance, the initial frame is assigned a \texttt{Serial Number} of 1, the subsequent frame is designated with a \texttt{Serial Number} of 2, and this pattern continues accordingly.

During the frames interpolation phase, the input text is reiterated prior to the generation of each interpolated frame, with the addition of a \texttt{Serial Number} to each. Unlike the key frame generation, the \texttt{Serial Number} following the first key frame is modified to reflect the total number of frames to be inserted plus 2, which accounts for the two key frames already established. Consequently, the \texttt{Serial Number} for the third frame is set to 2, the fourth to 3, and so on. This adjustment is made in anticipation of the frame reorganization step, where the third frame will be subsequently shifted to the second position in the sequence. 
\subsection{Redundant-Free Frame Interpolation}
After generating the key frames, a Recursive Interpolation process is employed to populate frames between each pair of key frames. A discriminative mechanism has been developed to enable the model to automatically omit the generation of redundant token patches during this process, thus accelerating the speed of interpolation.

\begin{figure}[htbp]
  \centering
\includegraphics[width=\linewidth]{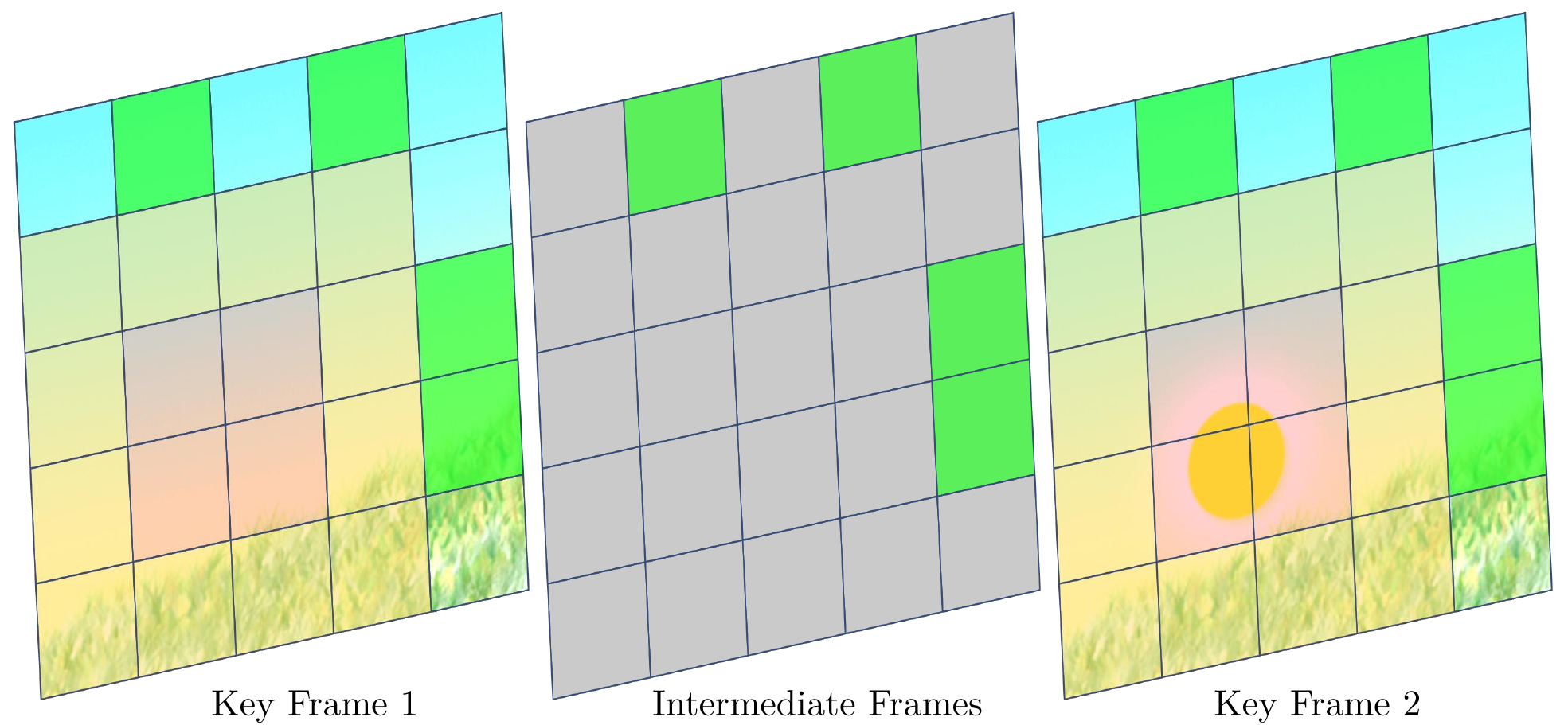}
  \caption{The Different Regions We Divide During the Interpolation Process. Red patches indicate the \textcolor{red}{Different Tokens}, orange patches denote regions of \textcolor{orange}{Unstable Tokens}, and green sections represent \textcolor{green}{Inheritable Tokens}.}
\label{fig:Redundant_Free}
\end{figure}

As illustrated in Figure~\ref{fig:Redundant_Free}, the initial step involves identifying sections that differ between the two key frames. 
In the figure, we have marked these tokens in red, which are defined as \textbf{\textcolor{red}{Different Tokens}}. These areas are inevitably subject to change within the intermediate frames. Consequently, our trained Interpolation model is deployed to generate the vision tokens for these identified positions. Encircling the red patches, there is an array of orange patches. The vision tokens within this proximity are considered as the  \textbf{\textcolor{orange}{Unstable Tokens}} due to their likelihood of alteration in the intermediate frames. Nevertheless, the trained Interpolation model is utilized to generate the vision tokens for these yellow patches as well, ensuring enhanced fault tolerance when interpolating the intermediate frames.

The rest of the vision tokens are categorized as \textbf{Stable Tokens}, considering their minimal propensity for alteration in the interpolated intermediate frames. A subset of these stable tokens (denoted by green sections in the figure) is randomly selected, and their values are documented. In the subsequent generation of intermediate frames, these documented values are directly applied to their respective positions, bypassing the need for recalculation by the model. 
We refer to the subset selected from the \textbf{Stable Tokens} as \textbf{\textcolor{Green}{Inheritable Tokens}}, indicating that the values corresponding to these tokens can be directly inherited from the key frames. The size of the subset is significantly correlated with the number of frames we plan to interpolate between the key frames. When inserting one key frame per second at a frame rate of 60, choosing 20\% of the stable tokens as inheritable tokens is a judicious choice.

This implies that within the intermediate frames shown in the figure, only the vision tokens corresponding to the gray areas necessitate generation via the interpolation model. On the other hand, the vision tokens associated with the green areas are directly derived from the key frames, markedly boosting the efficiency of generating intermediate frames.

%% file: sec/4_Experiment.tex
\section{Experiment}
To validate the performance of FlashVideo, in Section~\ref{sec:quant_res}, we quantitatively assess its text-to-video generation quality and efficiency. Moreover, in Section~\ref{sec:quali_res}, we showcase several videos produced by FlashVideo for qualitative evaluation. 
\subsection{Experimental Setups} 
\subsubsection{Datasets}
To evaluate the performance of FlashVideo, we employ three established benchmarks: UCF-101~\cite{karpathy2014large}, Kinetics-600~\cite{carreira2017quo} and BAIR~\cite{finn2016unsupervised}. UCF-101 consists of over 13,000 video clips across 101 action categories, offering a varied test bed for action-based video synthesis. Kinetics-600 expands this with around 500,000 clips in 600 categories, providing a broad spectrum of human activities for training. The BAIR robot pushing dataset includes over 44,000 sequences of robot-object interactions, valuable for models learning object manipulation. 
\subsubsection{Metrcis}
For quality assessment of the generated videos, we use Fréchet Video Distance (FVD)~\cite{unterthiner2018towards} to gauge content realism, Peak Signal-to-Noise Ratio (PSNR)~\cite{korhonen2012peak} for accuracy, Structural Similarity Index Measure (SSIM)~\cite{bakurov2022structural} for structural integrity, and Learned Perceptual Image Patch Similarity (LPIPS)~\cite{kettunen2019lpips} for perceptual likeness. Furthermore, to assess inference speed, we measure the number of frames generated per second during the inference process. We calculate the average values based on three separate trials.
\subsubsection{Implementation details}
In the realm of data preprocessing, we employ icetk~\cite{thudmicetk} as our tokenizer of choice, notable for its dual compatibility with both imagery and textual data, alongside the capability to integrate custom-defined special tokens seamlessly. For the evaluation of metrics, we harness the comprehensive framework referenced in ~\cite{hu2023metrics}, encompassing a spectrum of measures including FVD, PSNR, SSIM, and LPIPS, specifically tailored for the BAIR dataset analysis. In the context of model training, our infrastructure comprises eight A100 GPUs, each boasting a substantial 80GB of memory, to facilitate the rigorous training regime of our model. This training extends across 1000 epochs on the UCF-101 dataset, 500 epochs on the Kinetics dataset, and 800 epochs on the BAIR dataset, respectively. For optimization, Adam has been selected for its reliable performance.

\subsection{Quantitative Results}
\label{sec:quant_res}
\subsubsection{Video generation quality}
We have undertaken a comprehensive quantitative evaluation of our model across three distinct datasets. While metrics like FVD offer a measure for video generation tasks, the absence of a standardized protocol complicates direct comparisons. FVD readings are susceptible to various influences, including the resolution of the generated frames, cross-dataset training of the model, and the nature of the inputs, such as the inclusion of frames. For example, the CogVideo~\cite{hong2022cogvideo} model, with its hefty 9.4 billion parameters, underwent pre-training on an expansive corpus of 5.4 million captioned videos. Striving for a fair comparison, we adopted a balanced protocol, setting the resolution at $160x160$, initiating training from the ground up, and utilizing class-conditional inputs that comprise both text descriptions and the initial frame, thereby challenging the model to synthesize the ensuing frames. The findings from our experiments on the UCF-101, Kinetics600, and BAIR datasets are detailed in Tables~\ref{tab:UCF-101}, ~\ref{tab:k600}, and ~\ref{tab:bair}, respectively. On the UCF-101 dataset and Kinetics600 dataset, our model achieved FVD scores of 408 and 25.2, respectively. This experimental result indicates that although we are introducing a novel framework to the text-to-video generation task, it performs commendably when compared to some very mature model frameworks, such as autoregressive models (TATS-base~\cite{ge2022long}) and GAN models (DIGAN~\cite{yu2022generating}). On the BAIR dataset, we compared the generated frames using multiple metrics, and the results show that FlashVideo even exceeds the state-of-the-art (SOTA) in terms of LPIPS. This demonstrates that our model's generated frames have a high degree of congruence with the ground truth.
 
\begin{table}[h!]
\centering
\begin{tabular}{cccc}
\hline
\textbf{Method}                       & \textbf{Resolution}  & \textbf{Class}       & \textbf{FVD $\downarrow$}         \\ \hline
TGANv2~\cite{saito2020train}          & $128 \times 128$            & \checkmark                  & 1209                 \\
MoCoGAN-HD~\cite{tulyakov2018mocogan} & $128 \times 128$            &                    & 838                  \\
CogVideo~\cite{hong2022cogvideo}  & $480 \times 480$             & \checkmark                  & 626                  \\
DIGAN~\cite{yu2022generating}         & $128 \times 128$            &                    & 577                  \\
TATS-base~\cite{ge2022long}           & $128 \times 128$            &                    & 420                  \\
CCVS+StyleGAN~\cite{le2021ccvs}       & $128 \times 128$            & \checkmark                  & 386                  \\
Make-A-Video~\cite{singer2022make}    & $256\times 256$           & \checkmark                  & 367                  \\
TATS-base~\cite{ge2022long}           & $128 \times 128$            & \checkmark                  & 332                  \\ \hline
FlashVideo (ours)                      & \multicolumn{1}{c}{$160 \times 160$} & \multicolumn{1}{c}{\checkmark} & \multicolumn{1}{c}{408} \\ \hline
\end{tabular}
\caption{Video generation evaluation on UCF-101 dataset}
\label{tab:UCF-101}
\end{table}

\begin{table}[h!]
\centering
\begin{tabular}{cccc}
\hline
\multicolumn{1}{c}{\textbf{Method}}  & \textbf{Resolution}         & \multicolumn{1}{c}{\textbf{Class}} & \multicolumn{1}{c}{\textbf{FVD$\downarrow$}} \\ \hline
CogVideo~\cite{hong2022cogvideo}                             & $128 \times 128$                   &  \checkmark                                   & 109.2                            \\
CCVS~\cite{le2021ccvs}                                 & $128 \times 128$                   & \checkmark                                    & 55                               \\
Phenaki~\cite{villegas2022phenaki}                              & $128 \times 128$                   &                                    & 36.4                             \\
TrIVD-GAN-FP~\cite{luc2020transformation}                         & $128 \times 128$                   &                                    & 25.7                             \\
Video Diffusion~\cite{hoppe2022diffusion}                      & $64 \times 64$ &             \checkmark                        & 16.2                             \\ \hline
\multicolumn{1}{c}{FlashVideo (ours)} & \multicolumn{1}{c}{$160 \times 160$}        &      \checkmark                               & \multicolumn{1}{c}{25.2}             \\ \hline
\end{tabular}
\caption{Video generation evaluation on Kinetics-600 dataset}
\label{tab:k600}
\end{table}



\begin{table}[h]
\centering
\begin{tabularx}{\columnwidth}{@{}cXXXX@{}}
\hline
\textbf{Method} & \textbf{FVD$\downarrow$} & \textbf{PSNR$\uparrow$} & \textbf{SSIM$\uparrow$} & \textbf{LPIPS$\downarrow$} \\ \hline
CCVS~\cite{le2021ccvs} & 99 & - & 0.729 & - \\
MCVD~\cite{voleti2022mcvd} & 90 & 16.9 & 0.78 & - \\
MAGVIT~\cite{yu2023magvit} & 62 & 19.3 & 0.787 & 0.123 \\ \hline
FlashVideo(ours) & 83 & 17.1 & 0.741 & 0.098 \\ \hline
\end{tabularx}
\caption{Video generation evaluation on BAIR dataset}
\label{tab:bair}
\end{table}

\subsubsection{Video generation efficiency}
\begin{figure}[htbp]
  \centering
\includegraphics[width=\linewidth]{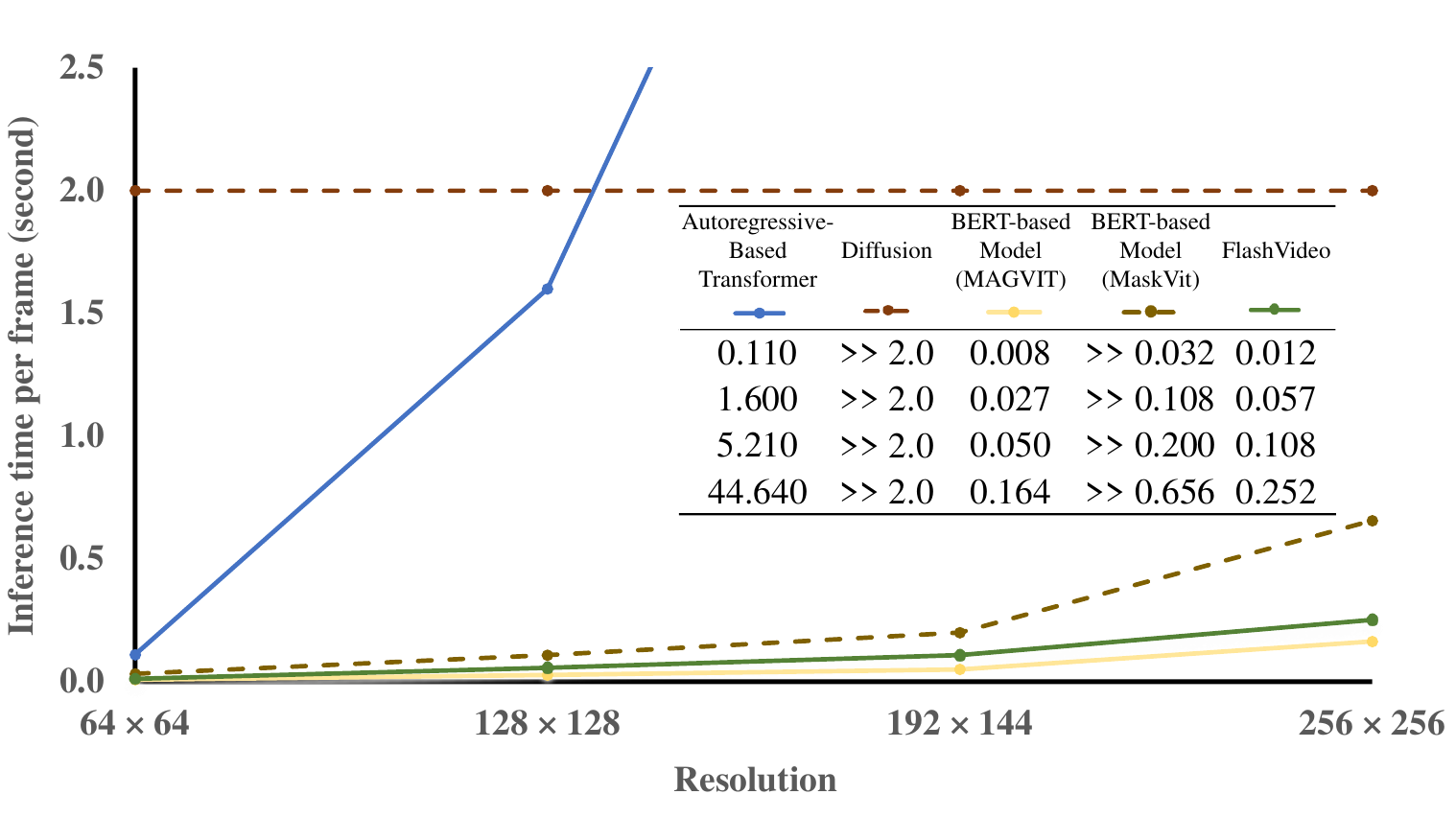}
  \caption{Comparison of Inference Efficiency. The diffusion model, denoted by the red dashed line, is referenced from the original paper which presents data only at a resolution of $64\times64$. Therefore, we employ the dashed line to convey that the actual inference time per frame for this model would be \textbf{no less than} the indicated value. As for the BERT-based MaskVit~\cite{gupta2022maskvit} model, the original publication does not provide explicit inference times. However, the MAGVIT~\cite{yu2023magvit} paper asserts that their approach is $4-16 \times$ faster than MaskViT. Based on this, our estimation is derived from the more conservative $4\times$ faster assertion. It should be noted that the true inference time for MaskViT is expected to exceed the indicated dashed line as well. Detailed data are recorded in the table embedded inside the figure. All data points have been benchmarked on a V100 GPU for consistency.}
\label{fig:inference_compare}
\end{figure}

To validate the key feature of our model, swift inference, we measured the average rate of frame generation at various resolutions. 
For our comparative analysis, we chose three widely recognized categories of video generation models as benchmarks: Video Diffusion~\cite{ho2022imagen} for diffusion models, TATS-base~\cite{ge2022long} for autoregressive-based transformer models, and BERT-based transformer models, specifically MAGVIT~\cite{yu2023magvit} and MaskVit~\cite{gupta2022maskvit}. The results are illustrated in Figure~\ref{fig:inference_compare}. 

The data compellingly illustrates that our FlashVideo model has remarkably optimized the time complexity from a quadratic $\mathcal{O}(L^2)$ to a linear $\mathcal{O}(L)$, as visually corroborated by the blue and green lines' comparison. When juxtaposed with the diffusion model, denoted by the red dashed line, FlashVideo's generation speed has escalated by roughly two orders of magnitude, a leap clearly depicted by the comparison between the red dashed and green lines. Furthermore, the study presents a nuanced comparison of FlashVideo against two eminent BERT-based transformer models, MAGVIT and MaskVit, showcasing our model's inference rate to be competitively situated between them, all within the same magnitude order—this is graphically represented by the proximity of the red line, yellow line, and brown dashed line.

\label{sec:quali_res}
\begin{figure*}[ht]
  \centering
  \includegraphics[width=\textwidth]{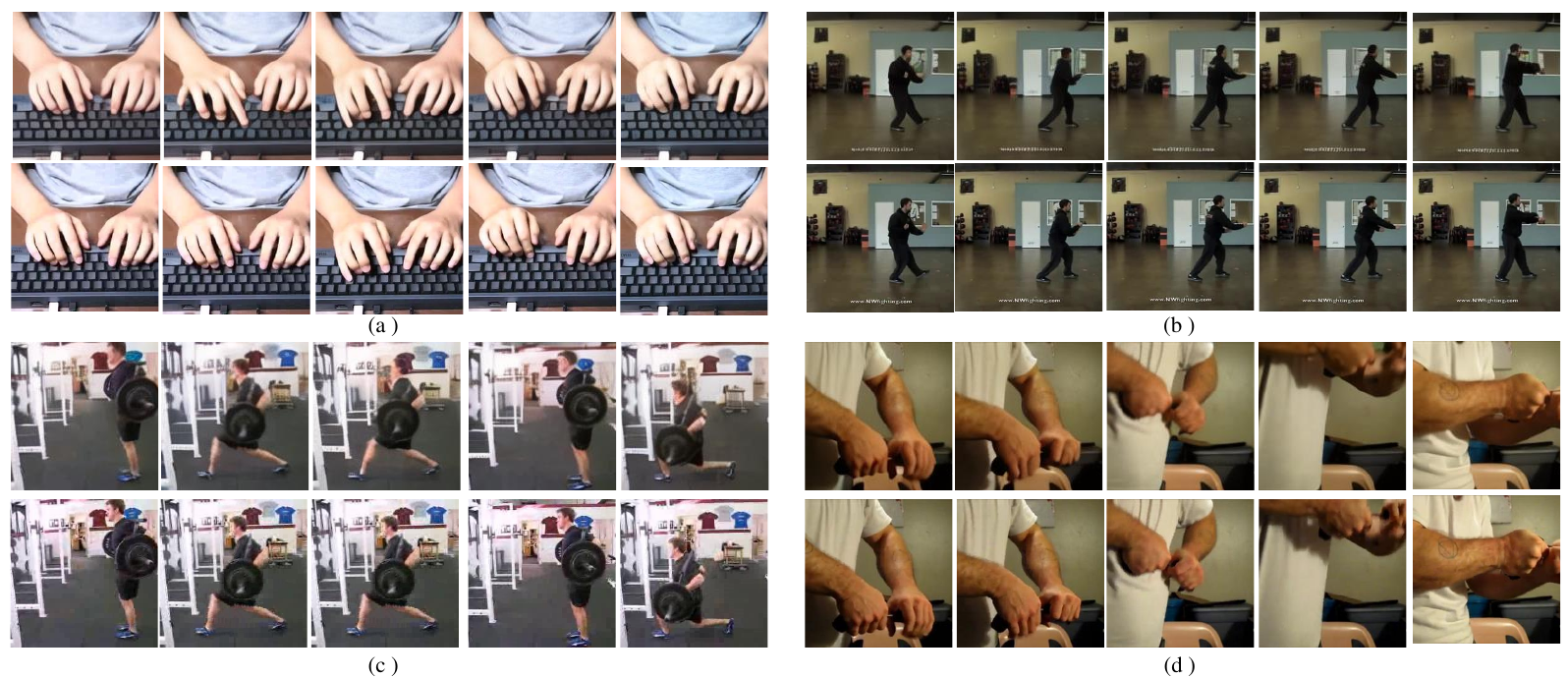}
  \caption{Qualitative evaluation. We juxtaposed the key frames generated by FlashVideo (Top row for each set) with their corresponding Groundtruth (Bottom row for each set). For each category, the initial input comprised the class label and the first 5 frames from the original video. (a) class label: \textit{Typing}, (b) class label: \textit{Tai Chi}, (c) class label: \textit{Lunges}, (d) class label: \textit{Bending metal}}
  \label{fig:Frames_figure}
\end{figure*}

\subsection{Qualitative Evaluation}
In this section, we delve into the capabilities of FlashVideo, showcasing its prowess in synthesizing video frames during the testing phase. We meticulously compare the generated frames against the ground truth from the original dataset, providing a qualitative analysis of the model's performance. Figure~\ref{fig:Frames_figure} offers a visual representation of this comparison, highlighting the efficacy of FlashVideo in replicating true-to-life motion and continuity across various activities.

The side-by-side comparison elucidates FlashVideo's nuanced understanding and recreation of complex motion dynamics. For the seemingly simplistic activities like \textit{Typing} (Figure a) and the fluid movements of \textit{Tai Chi} (Figure b), the model's output frames not only capture the rhythm and finesse but also mirror the precise posture and movement trajectory found in the ground truth. When tackling more intricate motions such as \textit{Lunges} (Figure c) and \textit{Bending Metal} (Figure d), FlashVideo demonstrates a robust capability to retain the core motion essence, even as it introduces unique elements that were not present in the initial frames. This distinct capability to generate frames that exhibit significant variations from the input signifies FlashVideo's advanced ability to understand and interpret complex activities. It underscores the model's potential to create not just a sequence of frames but a narrative of movement, providing insights into the sophistication of its underlying generative mechanisms.

%% file: sec/5_Conclusion.tex
\section{Conclusion}
Compared to current video generation models using GANs, transformer-based models, and diffusion models, FlashVideo successfully integrates the innovative RetNet architecture into this domain. Our experimental results demonstrate that FlashVideo not only competes with leading video generation models in terms of output quality but also sets a new standard in generation speed compared to autoregressive-based transformer models.  It surpasses diffusion models by two orders of magnitude and autoregressive transformer models by one order of magnitude in terms of inference efficiency, while achieving a comparable rate to BERT-based transformer models.  These accomplishments highlight the efficiency and effectiveness of FlashVideo, positioning it as a potential game-changer in video generation technology. Its successful adoption of RetNet opens new avenues for future advancements, setting a precedent for further innovations in efficient and high-quality video production.
